\documentclass[a4paper,fleqn]{cas-dc}

\usepackage[authoryear]{natbib}

\def\tsc#1{\csdef{#1}{\textsc{\lowercase{#1}}\xspace}}
\tsc{WGM}
\tsc{QE}
\usepackage{url}
\usepackage{graphicx}
\usepackage{multirow}
\usepackage[T1]{fontenc}
\usepackage{fancyhdr}
\usepackage{subcaption}
\usepackage{subfiles}
\usepackage{amsmath}
\usepackage{soul}
\usepackage{booktabs}
\usepackage{array}
\newcommand{\PreserveBackslash}[1]{\let\temp=\\#1\let\\=\temp}
\newcolumntype{C}[1]{>{\PreserveBackslash\centering}p{#1}}
\newcolumntype{R}[1]{>{\PreserveBackslash\raggedleft}p{#1}}
\newcolumntype{L}[1]{>{\PreserveBackslash\raggedright}p{#1}}
\definecolor{newgreen}{rgb}{0,0.56,0}

\begin{document}
\let\WriteBookmarks\relax
\def\floatpagepagefraction{1}
\def\textpagefraction{.001}

\shorttitle{DEPTWEET: A Typology for Social Media Texts to Detect Depression Severities}

\shortauthors{Kabir et al.}  

\title [mode = title]{DEPTWEET: A Typology for Social Media Texts to Detect Depression Severities}



\author[1]{Mohsinul Kabir}

\ead{mohsinulkabir@iut-dhaka.edu}
\credit{Conceptualization, Methodology, Software, Formal analysis, Investigation, Data Curation, Writing - Original Draft}

\author[1]{Tasnim Ahmed}
\ead{tasnimahmed@iut-dhaka.edu}
\credit{Methodology, Software, Formal analysis, Investigation, Writing - Original Draft}

\author[1]{Md. Bakhtiar Hasan}
\ead{bakhtiarhasan@iut-dhaka.edu}
\credit{Methodology, Formal analysis, Investigation, Writing - Original Draft}

\author[2]{Md Tahmid Rahman Laskar}
\ead{tahmedge@cse.yorku.ca}
\credit{Resources, Writing - Review \& Editing, Investigation}

\author[3]{Tarun Kumar Joarder}
\ead{tarun_psy@ru.ac.bd}
\credit{Supervision, Methodology, Validation}

\author[1]{Hasan Mahmud}
\ead{hasan@iut-dhaka.edu}
\credit{Writing - Review \& Editing, Resources, Supervision, Project administration, Funding acquisition}

\author[1]{Kamrul Hasan}
\ead{hasank@iut-dhaka.edu}
\credit{Writing - Review \& Editing, Resources, Supervision, Project administration, Funding acquisition}

\affiliation[1]{organization={Department of Computer Science and Engineering, Islamic University of Technology},
            addressline={Board Bazar}, 
            city={Gazipur},
            postcode={1704}, 
            state={Dhaka},
            country={Bangladesh}}

\affiliation[2]{organization={Department of Computer Science, York University},
            addressline={4700 Keele St}, 
            city={Toronto},
            postcode={M3J 1P3}, 
            state={Ontario},
            country={Canada}}

\affiliation[3]{organization={Department of Psychology, University of Rajshahi},
            addressline={Matihar}, 
            city={Rajshahi},
            postcode={6205}, 
            country={Bangladesh}}


%


%





\begin{abstract}
Mental health research through data-driven methods has been hindered by a lack of standard typology and scarcity of adequate data. In this study, we leverage the clinical articulation of depression to build a typology for social media texts for detecting the severity of depression. It emulates the standard clinical assessment procedure Diagnostic and Statistical Manual of Mental Disorders (DSM-5) and Patient Health Questionnaire (PHQ-9) to encompass subtle indications of depressive disorders from tweets. Along with the typology, we present a new dataset of 40191 tweets labeled by expert annotators. Each tweet is labeled as `non-depressed' or `depressed'. Moreover, three severity levels are considered for `depressed' tweets: (1) mild, (2) moderate, and (3) severe. An associated confidence score is provided with each label to validate the quality of annotation. We examine the quality of the dataset via representing summary statistics while setting strong baseline results using attention-based models like BERT and DistilBERT. Finally, we extensively address the limitations of the study to provide directions for further research.
\end{abstract}




\maketitle

\section{Introduction}\label{sec:intro}
Analyzing the presence of mood and psychological disorders through behavioral and linguistic cues from social media data remains a critical area of interdisciplinary research. In addition to these disorders, the last decade has seen exponentially increasing attempts to assess related symptomatology such as depressive disorders, self-harm, and severity of mental illness using non-clinical data \citep{Bucci2019TheDR}. Social media platforms and other online discussion forums have been particularly appealing to the research community for various research purposes (e.g., population-level mental health monitoring \citep{Conway2016SocialMB}, personal traits detection \citep{Marouf2020ComparativeAO}, cyberbullying spotting \citep{Bozyigit2021cyber}, etc.)  because of the massive scale of data. This massive data flow has resulted from increasing rates of internet access and people spontaneously sharing their suffering, pain, and struggle anonymously on these platforms \citep{Ofek2015SentimentAI}. Recognizing the early symptoms of depressive disorder through a person's language use can prevent many disastrous outcomes like self-harm, suicide, etc., and even help deploy effective treatment in proper time. Moreover, the outbreak of the COVID-19 pandemic is likely to have devastating impacts on the mental health of millions of individuals as lockdown in the affected areas has reported in high rises in the incident rates of mood disorder, including acute stress disorder, post-traumatic stress disorder, generalized anxiety disorder, and overall sub-clinical mental health deterioration \citep{singh_impact_2020}. The scope of mental health deterioration during the COVID-19 pandemic and the comprehensive nature of diagnosing depressive disorders have provided an unprecedented need to infer the mental states of individuals from all-inclusive resources. Recent studies have revealed that valuable insights into the impact of the pandemic on population-level mental health can be inferred from posts or comments on social media \citep{Low2020NaturalLP}.


A persistent challenge for the researchers specific to the mental health space is the need to: (a) establish a typology for text contents on social media to detect the severity of mental illness with clinical validation and robustness \citep{Ernala2019MethodologicalGI}, and (b) reliably apply this typology to obtain a sufficient sample size of high-quality data. Prior research has explored opportunities to capture mental health states from social media data using regular expressions to identify self-reported diagnosis or by using vectorization-driven methods to cluster activity patterns of users. However, deliberately relying on self-labeled data or unsupervised clustering leads to oversimplification and lacks clinical efficacy \citep{Ernala2019MethodologicalGI}. Practical exertion of mental health research includes identifying risky behaviors and providing timely interventions such as suicide prevention efforts adopted by Facebook \citep{vincent_facebook_2017}. The availability of high-quality, large-scale, annotated datasets addressing the severity of mental illness is one of the key elements for advancement on this front. Unfortunately, there are very few available datasets for depression severity which also lacks strong ground truths based on clinical validation \citep{tolentino_2018_dsm}. 

This study aims to contribute in this domain through (a) establishing a typology for social media contents (i.e., tweet text) built upon a psychological theory for detecting the severity of the mental condition of depressed individuals, (b) constructing a dataset named DEPTWEET\footnote{The DEPTWEET dataset is available at \href{https://github.com/mohsinulkabir14/DEPTWEET}{https://github.com/mohsinulkabir14/DEPTWEET}} containing around 40191 tweets with corresponding crowdsourced labels and confidence scores. The labeling typology of the dataset assigns a higher-level classification to each tweet, such as (1) Non-depressed, (2) Mildly Depressed, (3) Moderately Depressed, and (4) Severely Depressed. There is also an associated confidence score (between 0.5 and 1) for each label. 

The procedure used to assess the severity of depression in this study was based on a well-established clinical assessment method known as the Diagnostic and Statistical Manual of Mental Disorders, Fifth Edition (DSM-5) \citep{Arbanas2015DiagnosticAS}, and it was carried out under the supervision of two expert clinical psychologists. The DEPTWEET dataset contributes further high-quality data on attributes like none, mild, moderate or severe depression, adding to existing datasets on these and related attributes \citep{ahmed2021attention, mukhiya2020adaptation}, and provides the first dataset of this scale on depression severities to the best of our knowledge.  The approach utilized in this study can be adopted to generate high-quality mental health data from various platforms in future investigations. Moreover, given that the data was collected in the latter half of 2021, topic modeling on this dataset can provide useful insight into the impact of the COVID-19 pandemic on individuals' mental health.

The remaining sections of the paper are structured as follows: Section \ref{sec:related_works} and \ref{sec:measuring_severity} outlines the motivation and background of the DEPTWEET dataset. The data collection, quality control mechanisms, and the summary statistics of the data are described in Section \ref{sec:sourceData}. The baseline classification model for this dataset and evaluation metrics are presented in Section \ref{sec:experimental_design}. Section \ref{sec:result_discussion} discusses the classification results, potential sources of bias in the data, and the necessary aspects to consider while conducting additional research in this domain. Finally, Section \ref{sec:conclusionsAnd} draws a conclusion to the current study and discusses future directions.


\section{Related Work}\label{sec:related_works}

Computational linguistics techniques are very difficult to be opted as a complete substitute for in-person mental illness diagnosis, but the successful application of this domain in identifying the progress and level of depression of individuals in online therapy may provide clinicians with more insights, allowing them to apply interventions more effectively and efficiently. Studies analyzing web data, especially social media platforms, have piqued the interest of the research community due to their scope and deep entanglement in contemporary culture \citep{Fuchs2015CultureAE}. \citet{Coppersmith2014QuantifyingMH} made a prominent contribution in this domain by developing a procedure of extracting mental health data from social media. In their study, tweets were crawled from user profiles who publicly stated that they had been diagnosed with various mental illnesses on their Twitter feed. They mixed control samples from the general population (people who are not depressed) with the tweets of the self-reported diagnosed group. Additionally, they conducted an LIWC (Linguistic Inquiry Word Count) analysis to measure deviations of each disorder group from the control group. They focused on the analysis of four mental illnesses: Post-Traumatic Stress Disorder (PTSD), Depression, Bipolar Disorder, and Seasonal Affective Disorder (SAD), and proposed this novel method to gather data for a range of mental illnesses quickly and cheaply. Numerous studies later followed this approach to detect relevant mental health data for various mental illnesses. For example, The Computational Linguistics and Clinical Psychology (CLPsych) 2015 shared task \citep{Coppersmith2015CLPsych2S} collected self-reported data on Depression and PTSD. They further annotated the data with human annotators to remove jokes, quotes, etc., from the collected data. The shared task participants had three binary classification tasks- identify depression vs. control, identify PTSD vs. control, and identify depression vs. PTSD. These datasets were used in a variety of studies to discover patterns in the language use of users suffering from various mental illnesses \citep{Pedersen2015ScreeningTU, Coppersmith2016ExploratoryAO, Amir2017QuantifyingMH}. In particular, \citet{Resnik2015BeyondLE} conducted several topic modeling (supervised Latent Dirichlet Allocation (LDA), supervised anchor topic modeling, etc.) to differentiate the language usage of depressed and non-depressed individuals using the datasets of \citet{Coppersmith2014QuantifyingMH} and CLPSych Shared Task (2015).

Following a similar approach, \citet{Chen2018WhatAM} collected tweets from self-reported depressed users and investigated the potential of non-temporal and temporal measures of emotions over time to identify depression symptoms from their tweets by detecting eight basic emotions (e.g. anger, fear, etc.). Additionally, classifiers were built to label Twitter users as either depressed or non-depressed (control) groups calculating the strength scores based on the intensity of each emotion and a time series analysis of each user. Among other social medias, \citet{Xianyun2016weibo} explored sleep complaints on Sina Weibo (a Chinese microblogging website) to discover users' diurnal activity patterns and gain insight into the mental health of insomniacs. Twitter data on mental health had also been collected, with specific Twitter campaigns being targeted. For instance, \citet{Jamil2017MonitoringTF} prepared a dataset from the users who participated in the \#BellLetsTalk 2015 campaign that was inaugurated to promote awareness about mental health issues. They collected public tweets from 25362 Canadian users and built a user-level classifier to detect at-risk users and a tweet-level classifier to predict symptoms of depression in tweets. From this campaign, they came across only 5\% tweets that talk about depression and 95\% non-depressed tweets. While these methods can extract large volumes of data for a low cost, they do not ensure a sufficient sample of interest and have inevitably resulted in a low number of positive samples (mental-health related data). 

Several previous studies have investigated the use of clinical methodologies along with data mining tools to extract depression symptoms from diverse sources. \citet{yazdavar2017semi} created a lexicon of depression symptoms based on the nine disorders described in the clinically established Patient Health Questionnaire (PHQ-9) and utilized this to find symptoms of depression in tweets from users with self-reported depressive symptoms in their Twitter profile. They also developed a statistical model to categorize and monitor depressive symptoms for continuous temporal analysis of an individual's tweets. In a similar study, \citet{mukhiya2020adaptation} proposed an open set of depression word embeddings that extracts depression symptoms from patient-authored text data based on PHQ-9 to deliver personalized intervention to people with symptoms of depression. \citet{Yadav2020IdentifyingDS} utilized the nine symptom classes of the PHQ-9 questionnaire to manually annotate the tweets collected from 205 self-reported depression diagnosed users. Their proposed framework took into consideration the figurative language (metaphor, sarcasm etc) wired in the communication of depressive users on Twitter. \citet{ahmed2021attention} extracted depression symptoms in patient authored text in a similar fashion with PHQ-9 questionnaire but used an attention-based in-depth entropy active learning to annotate the unlabeled texts automatically. Their mechanism increased the trainable instances of mental health data using a semantic clustering mechanism with to reduce the data annotation task. Another mental health tool used by psychiatrists, namely the Diagnostic and Statistical Manual of Mental Disorders (DSM-5), has also been used to categorize mental disorders from social media content. \citet{Gaur2018LetMT} developed an approach to map subreddits into DSM-5 categories. They created a lexicon from various subreddit posts by extracting n-grams and topics using LDA and mapped this lexicon with DSM-5 lexicon created by available medical knowledge bases (ICD-10\footnote{https://bioportal.bioontology.org/ontologies/ICD10}, SNOMED-CT\footnote{http://bioportal.bioontology.org/ontologies/SNOMEDCT}, DataMed\footnote{https://datamed.org/}). Their approach attempted to connect a patient on social media platforms such as Reddit to appropriate mental health resources and to provide web-based intervention. \citet{CavazosRehg2016ACA} investigated the most common themes of depression-related chatter on Twitter that corresponded to the DSM-5 symptoms for major depressive disorder. While these methods may have clinical validity, most studies that use them lack sufficient ground truth data due to the absence of a thorough annotation procedure.

Very few studies have investigated predicting the severity of depression based on users' language usage on web platforms. \citet{Choudhury2013SocialMA} proposed a metric named social media depression index (SMDI) using a probabilistic model to help characterize the levels of depression in the population level. This probabilistic model is an SVM classifier that can predict whether or not a Twitter post contains symptoms of depression. To construct and train this model, they collected data using crowdsourcing technique and derived various linguistic and network features (e.g., number of followers) from tweets of individuals suffering from clinical depression, which was measured using the CES-D (Center for Epidemiologic Studies Depression Scale) screening test \citep{Radloff1977TheCS}. \citet{Schwartz2014TowardsAC} attempted to predict and characterize the severity of depression based on people's Facebook language use. They gathered survey responses and Facebook posts from 28749 Facebook users and trained a classification model to predict depression symptoms using n-grams, linguistic behavior, and LDA topics. They tried to quantify the seasonal changes in depression symptoms based on social media posts and discovered that symptoms increase from summer to winter. These approaches had the potential to generate a large dataset with good quality data if they were developed in partnership with expert psychologists and domain experts. 

While previous research has made significant progress toward automatic depression assessment tools based on social media, some limitations have been identified through critical evaluation. Most previous works have relied on self-reported depressed user profiles when it comes to data extraction. While this is an inexpensive way to gather a massive scale of data, it doesn't guarantee enough samples with depressive symptoms without manual intervention. Also, this approach might lack enough clinical validation to extract depression symptoms. Studies that leveraged clinical assessment tools to extract data, such as the PHQ-9 or DSM-5, lacked supervision from domain experts and mostly annotated their data in an automated manner, such as using unsupervised topic modeling or clustering techniques. Moreover, only a few studies have investigated how to collect data on different depression severities with sufficient clinical efficacy. The existing datasets only concentrate on binary detection of whether a particular tweet manifests depression or not, the severity level of which is mostly ignored. This might lead to models competent enough in detecting subtle cues of depression turn a blind eye towards them. A dataset containing sufficient samples to train large models with strong ground truth labels depicting the severity of depression can go a long way to alleviate these issues.



\section{Measuring Severity of Depression}\label{sec:measuring_severity}

In the current study, a user posting a tweet on social networking site Twitter is considered to be depressed if the tweet depicts behaviors portraying symptoms of depression. Such a tweet may not necessarily be complete, contain well-structured sentences, or even grammatically correct, making the task even more difficult.

According to the Diagnostic and Statistical Manual of Mental Disorders (DSM), clinical depression can be diagnosed considering the existence of a set of symptoms over a substantial amount of time \citep{yazdavar2017semi}. Incorporating this idea, the Patient Health Questionnaire (PHQ-9) \citep{kroenke2001phq} provides a set of questionnaires, which is widely used to screen, diagnose and measure the severity of depression. Using this set of questionnaires, nine distinct symptoms related to different disorders, such as lack of interest, eating disorder, etc., can be extracted (\tablename~\ref{tab:seed_terms_and_keywords}).

\begin{table*}[t]
    \centering
    \caption{Sample tweets, seed terms and final keywords list for each symptom of PHQ-9 Questionnaire}
    \label{tab:seed_terms_and_keywords}
    \begin{tabular}{L{3cm} L{3.8cm} L{2.7cm} L{4cm}} 
    \toprule
     \textbf{PHQ-9 Symptoms} & \textbf{Sample Tweet} & \textbf{Seed Terms} & \textbf{Final Keyword List}\\ 
   \midrule
    Lack of interest (S1) & Am I depressed or am I just bored? Apathy and irony, postmodern anxiety & disinterest & involved, occupied, pessimism, reversion, absorbed, lifelessness, bored, enthusiasm, engrossed, worried, apathy. 
  \\
    Feeling Down (S2) & High functioning depression, I can’t fester in my misery but i’m fuckin miserable & hopeless,  depressed  & dejected, dismayed, dispirited, demoralized, grimmed, misery, 
    grim, downhearted, low-spirited, bleak, desperate, lost, frustrated.
   \\ 
    Sleep Disorder (S3) & forcing myself up now so I'm not awake when the power goes off much later, lol &  awake, sleep & nap, restless, awake, whole night, bedtime. 
 \\ 
    Lack of Energy (S4) & I’m so exhausted and I still have work 9-5 and then red rocks day three & tired, energy & weary, fatigue, fag, fag out, overtire, overfatigued, burned-out, burnt-out, exhausted, dog-tired, washed-out, drained, whacked.
 \\ 
    Eating Disorder (S5) & another saturday night where i’m too depressed to sleep after overeating....i am extremely bored of this life & appetite, overeating & aversion, distaste, loathing, malformed, bulimic, puffy, starve, fat
\\
    Low Self-estemm (S6) & I got on the scale today and I am disgusted. Like utterly disgusted. Depression really beat my ass and had me slacking & loser, failure & loser, relapse, downfall, ruined, flop, dead-duck, disappointment,
    achiever, misfire, underdog, falling-apart, disgusted 
 \\ 
    Concentration Problems (S7) & Whenever it gets close to my bday I always go through some type of cleansing/depression.. Scattered focus... & concentrate, focus & immersed, decentralize, deconcentrate, scattered, dispersed, unsettled, focus 
  \\
    Hyper/Lower Activity (S8) & I spend hours of my day staring at screens, immobile. Why am I depressed??? & moving, immobile, restless & discontent, ungratified, unsatisfied, stand-still, refrained, immobile \\
   
    Suicidal Thoughts (S9) & I know that I can't undo The self-destruction, the damage I've done & dead, hurt, suicide & trauma, harm, suffering, anguish, hemorrhage, penetrating-trauma,
    torment, agony, excruciate, damaged, gag, suffocate, self-destruction 
   \\ \bottomrule
    \end{tabular}
    
\end{table*}

The frequency of these symptoms can help classify the severity of depression as none, mild, moderate, and severe conditions. This approach is called Clinical Symptom Elicitation Process (CSEP) \citep{world1993icd}. In the current study, this was further extended using the mood scale provided by BipolarUK\footnote{\href{https://www.bipolaruk.org/faqs/mood-scale}{https://www.bipolaruk.org/faqs/mood-scale}} to identify the characteristics related to different levels of depression. The following characteristics were then verified by the collaborator psychologists and used to detect the level of depression from the user tweets:


\subsection{Non-depressed Tweets}
A tweet can be labelled as a non-depressed tweet if it expresses a person's joy or delight, or makes a generalized statement about depression that does not reflect the person's own mental state, expresses casual tiredness or sadness (For example, sadness due to the defeat of their favorite sports team), or expresses temporary hopelessness. It can also convey any other emotion except for depression.
\subsection{Mildly Depressed Tweets}
A tweet that expresses hopelessness or a feeling of disinterest that persists for a while can be labeled as a mildly depressed tweet. A mildly depressed tweet may contain symptoms of hopelessness, feelings of guilt or despair, difficulties concentrating at work, a loss of interest in activities, a sudden disinterest in socializing, a lack of motivation, insomnia, weight changes, daytime sleepiness and fatigue, appetite changes, and reckless behavior (such as, alcohol and drug abuse).
\subsection{Moderately Depressed Tweets}
Moderate depression has symptoms similar to mild depression. The differentiating factor is that the severity of symptoms hampers activities related to home and work. Tweets may contain symptoms of increased sensitivities, feeling of worthlessness, reduced productivity, problems with self-esteem, excessive worrying.
\subsection{Severely Depressed Tweets}
The symptoms of this category are more noticeable and life threatening.  They contain delusions, feeling of near-unconsciousness or insensibility, hallucinations, suicidal thoughts, or behaviors.

\section{The DEPTWEET Dataset}\label{sec:sourceData}

\begin{figure*}[t]
    \centering
    \includegraphics[width=\textwidth]{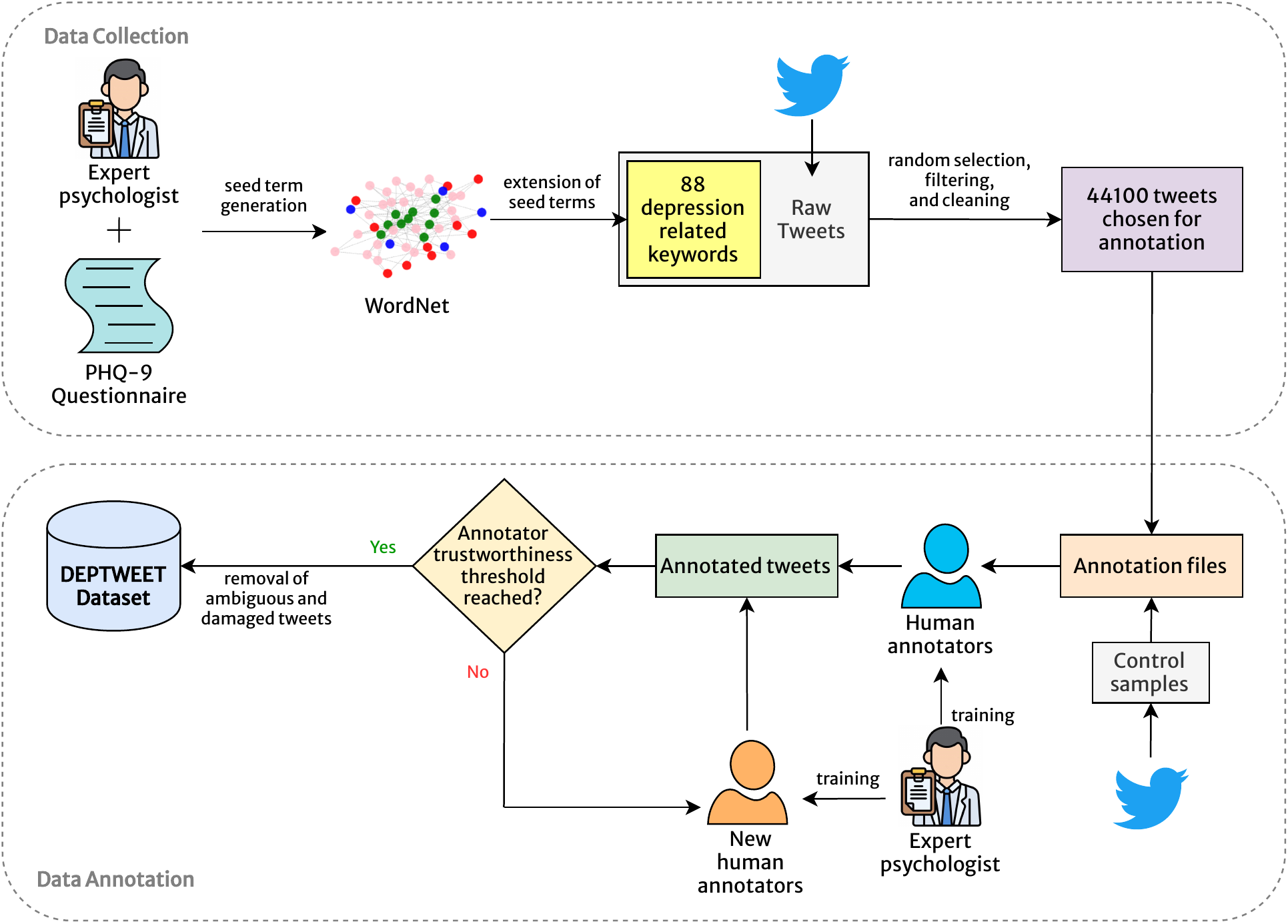}
    \caption{Overview of the dataset creation process}
    \label{fig:overview}
\end{figure*}

In this section, the complete methodology of constructing the DEPTWEET dataset and the summary statistics of the data is discussed extensively. TWINT\footnote{\href{https://github.com/twintproject/twint}{https://github.com/twintproject/twint}} was used to collect tweets from Twitter for this study. The collected tweets went through a preliminary screening process before being distributed to the annotators. The annotation job was carefully observed and regulated in order to maintain the high quality of the data. An overview of the data collection and annotation procedure is displayed in \figurename~\ref{fig:overview}. Below, we first present how we collected the data. Then, the data annotation process is demonstrated in detail. Finally, we discuss the properties of the  dataset.

\subsection{Data Collection}

Seed terms were generated from the keywords extracted from each of the symptoms of PHQ-9 questionnaire by collaborating with two professional psychologists. This is a commonly used procedure employed in many previous studies \citep{yazdavar2017semi,mukhiya2020adaptation,ahmed2021attention}. After seed terms generation, they were then extended using WordNet \citep{miller1995wordnet}. It is a well-known lexical database developed by Princeton University that links words into semantic relations, including synonyms, hyponyms, meronyms, and antonyms. Each category of words is maintained according to their parts of speech, i.e. nouns, verbs, adjectives, and adverbs in the database and the synonyms are grouped into synsets. Words that are in the same synset are synonymous and interlinked using conceptual-semantic and lexical relations. There are several other methods used in different studies \citep{mukhiya2020adaptation,yazdavar2017semi} such as \textit{Universal Sentence Encoding (USE)} \citep{cer2018universal}, \textit{Global vector representation (GloVe)} \cite{pennington2014glove}, \textit{Big Huge Thesaurus} \citep{watson2007big}, etc. In the evaluation shown by \citet{mukhiya2020adaptation}, WordNet performs significantly better in extracting symptoms from patient-authored text compared to other methods. For this study, the seed terms for each questionnaire of PHQ-9 were extended by WordNet, and the extended terms were handpicked afterward by the psychologist collaborators. After several rounds of filtration, a final lexicon list containing 88 depression-related keywords categorized into nine different clinical depression symptoms of PHQ-9 was prepared, which are likely to appear in the tweets of individuals suffering from different severities of depression. \tablename~\ref{tab:seed_terms_and_keywords} illustrates samples of anonymized tweets, seed terms, final keywords list extended by WordNet and their associated symptoms in PHQ-9. Based on the final keyword list, a total of 344657 tweets were collected.

\subsection{Data Annotation}
Several data annotation techniques can be applied to determine the class label for the sample tweets. Since the number of classes is known beforehand, one intuitive approach can be creating vector representations of the tweets using Word2Vec \citep{mikolov2013efficient}, GloVe \citep{pennington2014glove}, fastText \citep{piotr2017enriching}, etc. and then using unsupervised clustering to find the optimal distribution of the samples into different clusters. However, such approaches lack human input who can understand the subtle nuances of tweets to identify different levels of severity, resulting in a poorly annotated dataset \citep{Ernala2019MethodologicalGI}. To ensure clinical accuracy, annotators, trained by expert psychologists, were employed to perform dataset annotation.

From the collected samples, tweets that were posted in English were only preserved for annotation. Tweets with less than eight words were discarded as they might not contain enough context. Any tweets containing mentions (@) or hashtags (\#), as well as retweets, were also discarded since they could violate the privacy of the users mentioned. Finally, 44100 tweets were randomly chosen from the remaining tweets for annotation.

\subsubsection{Annotator Recruitment}

The annotation job was done by recruiting participants who were fluent in English and had a previous experience of text assessment. The annotator pool consisted of 111 crowdworkers, and they were pre-screened for eligibility using two online sessions. Initially, 90 annotators were selected randomly for the annotation job after pre-screening. Each annotator received \$20 for participating in the study. The task of the annotators was to label the tweets as one of the four classes, i.e., non-depressed, mildly depressed, moderately depressed, and severely depressed tweets. The annotators were briefed through 2 long online sessions under the supervision of the collaborator psychologists about the classification and were also provided with a detailed document on the severity classes. Each annotator was given a datafile with only two columns: (1) tweet texts and (2) possible label suggestions (0: non-depressed, 1: mild, 2: moderate, 3: severe) and was asked to determine the tweet's possible class label.

The inherent subtlety and ambiguity of the attributes covered in this dataset makes the annotation procedure an unavoidably difficult process. Each annotator may have a unique perspective on the nuance of the context presented in tweets, as well as a unique perception of the severity of the depression. Annotators were asked to avoid personal bias while labeling the tweets and strictly follow the guidelines provided to them to classify the text. Each tweet was annotated at least three times. The final label of a tweet was determined by majority voting of the labels provided by the three annotators. Tweets with different labels from all the three annotators were discarded because of too many disagreement. Final labels of the dataset were established with a confidence score to reflect the disagreement of the annotator because of reasonable difference of opinion. 

\subsubsection{Annotation Job Refinement}

Though it was ensured that annotators' disagreement reflected a genuine difference of opinion, a means of quality control was required to prevent annotators' inattention, or misunderstanding of context. The quality control mechanism used by \citet{price2020six} was followed in this study. This mechanism aimed to reduce the number of `bad' annotators, those who either did not correctly understand the task or annotated the datafiles too recklessly, without giving proper attention. As part of the quality control, a set of `control samples' was collated with the actual data sample, for which the correct labels were manually established. Annotators encountered one control sample per batch of fifty tweets without knowing which of the tweets was the control sample. The running accuracy of these control samples was defined as annotator's `trustworthiness score (T)'. The threshold trustworthiness score for this study was set to be at least 90\%. If an annotator dropped below this level, all of their annotations were discarded, and the annotator was removed from the annotator pool. Afterwards, another annotator from the pool was assigned to re-annotate those data samples.

A total of $900$ control samples were added for quality control with the previously chosen $44100$ data samples. To generate datafiles for the annotators, the actual dataset containing 44100 samples were divided into 30 parts, each part containing $(44100/30) = 1470$ samples. For every $49$ tweets in these $1470$ samples, one unique control sample was added at a random position. The control samples were from the \textit{non-depressed} category and were limited to only obvious and conclusive instances of attributes. Thus, one would fail on these control samples only if they had an incorrect comprehension of the attributes of the class labels or were too reckless while annotating. The tweet ID of the control samples were also tracked. Following this method, 30 datafiles were created containing (1470 data samples + 30 control samples) = 1500 tweets each. Each datafile consisted of two columns: one having tweet texts, and another empty column for annotator label. All the other data columns were kept hidden from the annotators. The metadata related to the datafile creation procedure is summarized in \tablename~\ref{tab:scraped_tweet_metadata}. To annotate these datafiles, ninety annotators were divided into three groups, each with thirty annotators. Each datafile was given to three different annotators from three different groups. Before partitioning, the data samples were randomized so that no two data files contained identical tweets in the same order. Once the annotation process was finished, all the datafiles were merged and the control samples were removed from the dataset.

\begin{table}[t]
\centering
\caption{Metadata about the datafiles created for annotation}\label{tab:scraped_tweet_metadata}
\begin{tabular}{l c}
\toprule \textbf{Type} & \textbf{Count}  \\ \midrule
Number of tweets collected & 344657 \\
Tweets chosen for annotation & 44100\\
Total datafiles created & 30\\
Data samples in each datafile & 1470\\
Control samples per datafile & 30\\
Total tweets per datafile & 1500\\
\bottomrule
\end{tabular}

\end{table}

\subsection{Dataset Properties \& Analysis}\label{sec:theDEPTWEET}

From the $44100$ tweets considered for annotation, $1399$ data samples were removed from the dataset because they were damaged (i.e., tweet text or tweet ID was changed) during the annotation process, and $2510$ data samples were discarded due to annotator disagreement, as they received three different labels from three different annotators. The final dataset comprises a total of 40191 tweets along with their \textit{tweet\_id}, \textit{replies\_count}, \textit{retweets\_count}, \textit{likes\_count}, \textit{target}, \textit{label} and \textit{confidence\_score}. The label for each tweet was determined based on the aggregation of the labels provided by different annotators. If at least two of the three annotators agreed on the label of a tweet, the matched annotation was accepted as the final label. Tweets that had three different annotations from three annotators, were discarded and saved in a separate datafile\footnote{Further annotation is required to achieve a class label for these samples and were left because of budget and time constraint.}. The corresponding confidence score for each label was determined by an weighted average of the annotator's `trustworthiness score'. Confidence Score for a particular label of a tweet sample can be written as: 
 \begin{align}
     \text{Confidence Score} (C) = \frac{\sum T_i}{T}
 \end{align}
 
where $T_i$ denotes trustworthiness of $i^\text{th}$ annotator whose annotations matched and $T$ denotes sum of the trustworthiness score of all the annotators who annotated the tweet.

To demonstrate this process, consider a tweet sample annotated by three annotators $A$, $B$, and $C$ having trustworthiness scores $T_A = 0.90$, $T_B = 0.93$, and $T_C = 1.00$. If the annotated label of annotators $A$ and $B$ matches, then the confidence score of the label will be $(T_A + T_B) / T$, where $T$ is the sum of the trustworthiness score of the three annotators. In this case, the confidence score for the label of the particular tweet would be $0.647$.

\begin{table}[h]
\centering
\caption{Percentage of Data Samples for Each Class}\label{tab:dataset_class_dist}
\begin{tabular}{l   l}
\toprule \textbf{Class} & \textbf{Proportion}  \\ \midrule
Non-depressed   & 80.62\% \\
Mild    & 13.04\%\\
Moderate    & 4.5\%\\
Severe  & 1.84\%\\
\bottomrule
\end{tabular}
\end{table}
 
Manual analysis was performed in two stages of the study to gain insights into the dataset: (i) while randomly choosing data samples for annotation, and (ii) during the initial iterations of the annotation job. The proportion of classes shown in \tablename~\ref{tab:dataset_class_dist} indicates that the \textit{non-depressed} samples outnumber the other classes by a wide margin. Though all the data samples were scraped based on the keywords related to different severity levels of depression and the control samples were removed prior to the final preparation of the dataset, the number of data samples for different severities of depression is inevitably low. This class imbalance represents an important characteristic in the identification of various depressive disorders on social media. The final class proportions roughly represent the percentage of similar attributes in similar live contexts. 

\begin{figure}[h]
    \centering
    \includegraphics[width=0.5\textwidth]{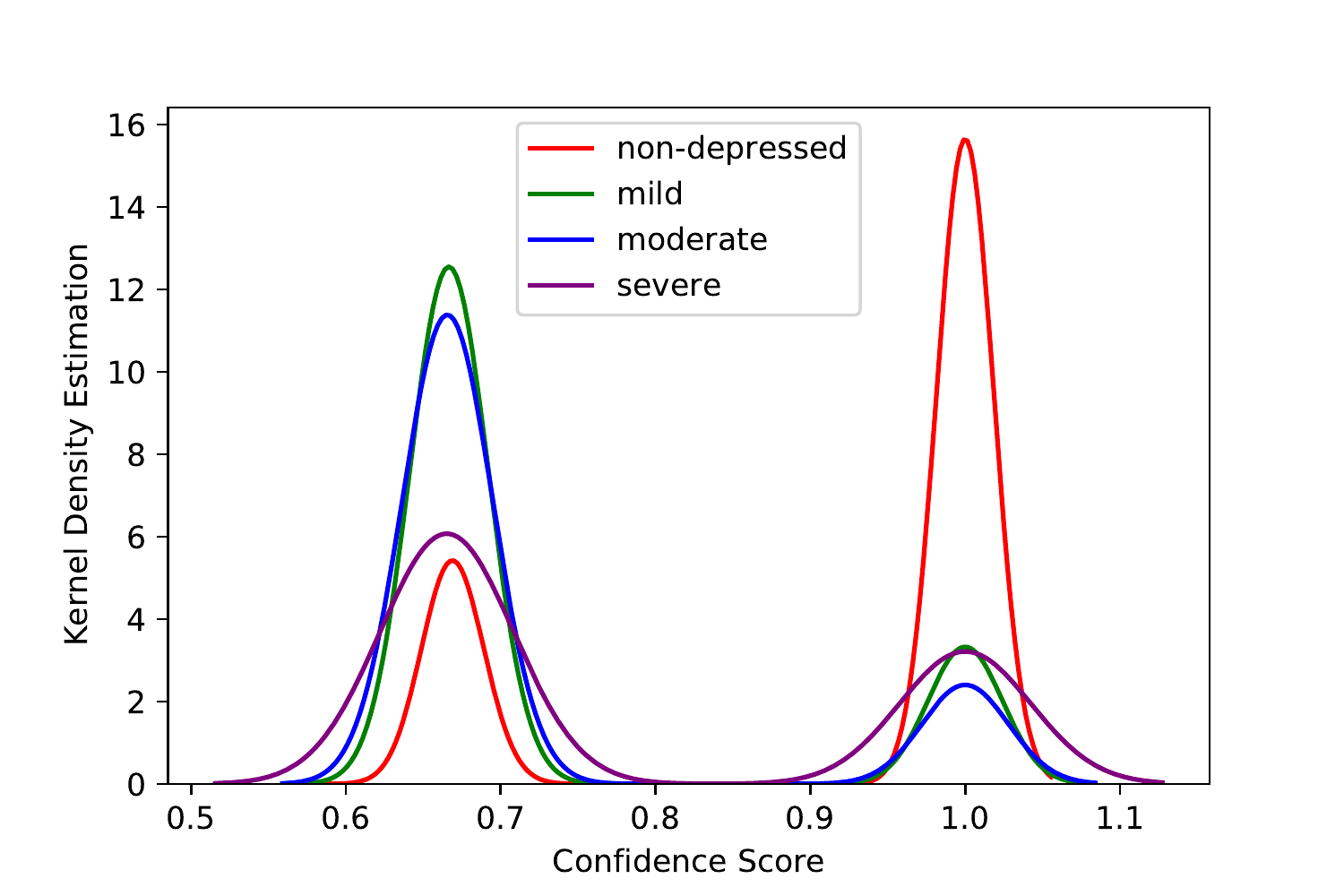}
    \caption{Kernel density estimation of confidence scores for each class}
    \label{fig:kde}
\end{figure}

Generally, the overall positive content shared in social media outnumbers the negative content. This is because people usually show their positive, friendly side over social media and tend to talk less about their struggles \citep{vermeulen2018smiling}. To mitigate this problem, previous studies depended on self-labeled data for collating large and balanced datasets on different mental disorders \citep{kim2020deep, Low2020NaturalLP}. However, depending only on self-labeled data to understand mental health from personal levels and measure the severity of the condition is not feasible without the intervention from expert psychologists. But considering the lack of resources in the mental health sector, only relying on psychologists can be time-consuming and expensive. As a result, in this study, crowdsourcing supervised by psychologists was opted to obtain high-quality data on different depression severities.

Despite the measures undertaken to ensure the quality of the dataset, the method of annotation warrants a certain level of noise. This results in different yet rational interpretations of the same tweet. The kernel density estimation of the confidence scores portrayed in \figurename~\ref{fig:kde} indicates that there was reasonable agreement among the annotators on deciding the class label of the \textit{non-depressed} and \textit{severe} classes. While these two classes lies on two different polarities of attributes, the subtle nuances of the \textit{mild} and \textit{moderate} classes allowed for rational disagreement among the annotators, which is evident from the high concentration of probability density for \textit{mild} and \textit{moderate} classes between 0.6 and 0.7 in \figurename~\ref{fig:kde}. This may be attributed not only to the lack of apprehension or awareness of the annotator, but also on the subjectivity of the topic at hand. It highlights the difficulty of using typical reliability metrics such as Inter-Rater Reliability (IRR), which calculates the level of agreement between two or more annotators. More sophisticated metrics like Fleiss' Kappa \citep{fleiss2013statistical} can be applied in this scenario since the sample tweets were distributed randomly among the annotators and each annotator chose from one of the four mutually exclusive labels to indicate the severity of depression per tweet \citep{gwet2014handbook, leard2019fleiss}. However, Fleiss' Kappa assumes that the disagreement among the annotators on the same sample reduces the reliability of the dataset. Considering the subjective nature of the severity of depression detected by different annotators, that might not be the case \citep{salminen2018inter}. In spite of that, Fleiss' Kappa was calculated to get an understanding of the overall agreement of the annotators in this study. The value of Fleiss' Kappa ranges from -1 (indicating no observed agreement) to +1 (indicating a perfect agreement) \citep{leard2019fleiss}. Here, a value less than 0.20 indicates a poor agreement, 0.21 to 0.40 indicates a fair agreement, 0.41 to 0.60 indicates moderate agreement, 0.61 to 0.80 indicates substantial agreement and 0.81 to 1 indicates a near perfect agreement among the annotators.

\begin{table}[h]
    \centering
    \caption{Fleiss' Kappa per class}
    \label{tab:fleiss}
    \begin{tabular}{l c}
        \toprule
        \textbf{Class} & \textbf{Fleiss' Kappa}\\
        \midrule
        Non-depressed & 0.44\\
        Mild & 0.27\\
        Moderate & 0.30\\
        Severe & 0.45\\
        \midrule
        Overall & 0.36\\
        \bottomrule
    \end{tabular}
    
\end{table}

As reported in \tablename~\ref{tab:fleiss}, the Fleiss' Kappa for the \textit{non-depressed} and \textit{severe} classes show a moderate agreement among the annotators. This can be explained considering the extreme nature of these two classes as they tend to be the polar opposite of each other. On the other hand, a fair agreement in \textit{mild} and \textit{moderate} classes highlight the intricate relationship among these two classes and the difficulty in identifying the subtle cues to differentiate them, even for the humans. However, despite the subjective nature of the severity of depression, an overall fair agreement provides indication of the quality of the annotation, and the dataset in general.

\section{Experimental Design}\label{sec:experimental_design}
The choice of baseline models and evaluation metrics for this study are discussed in this section.

\subsection{Baseline Models Selection} \label{ref:models}
One of the main challenges in language related tasks comes from the use of homonyms and synonyms as well as different kinds of ambiguity in sentences such as, lexical, semantic, and syntactic ambiguity. Another challenging task for a model is to extract context from various domain specific language. Empirical studies have shown that rule-based methods and traditional machine learning-based methods fail to overcome these complexities by understanding the inherent meaning of the sentences \citep{kansara2020comparison, gonzalez2020comparing}. Multilingualism is another challenge with classic machine learning techniques \citep{gonzalez2020comparing}. Rules for a specific language can be formed, but alphabet and even the sentence structure can differ from one language to other, requiring the development of new rules. Most of these aforementioned shortcomings are alleviated by transformer \citep{vaswani2017attention} based architectures that use attention mechanism to capture bi-directional context and also capable of handling larger datasets than traditional machine learning-based architectures.
Considering these issues, a series of baseline models were chosen to evaluate the proposed dataset, namely Support Vector Machine (SVM) \citep{Cortes2004SupportVectorN}, Bidirectional LSTM (BiLSTM) \citep{Schuster1997BidirectionalRN}, BERT \citep{devlin2018bert} and DistilBERT \citep{sanh2019distilbert}. Bidirectional LSTM (BiLSTM) was selected as it is a widely used recurrent neural network based on deep learning architecture, Support Vector Machine (SVM)  as a classical machine learning model, and BERT and DistilBERT as two transformer-based models to evaluate the dataset. 
While the word embedding of SVM and BiLSTM models rely on choice, both BERT and DistilBERT are pre-trained using a large amount of data from English Wikipedia\footnote{\href{www.wikipedia.org}{www.wikipedia.org}} and Toronto Book Corpus \citep{zhu2015aligning}. The pre-training is generic enough to be fine-tuned for downstream tasks such as sequence classification, named entity recognition, natural language inference, etc.

Reasons for choosing these models can be summarized as follows:

\begin{itemize}
    
    \item  A diverse set of classifiers are chosen as baseline models to evaluate the validity of the dataset. SVM has already been used by \citet{Choudhury2013SocialMA} to create a probabilistic model to predict severity of depression from tweets. BiLSTM is a sequence processing model that calculates the input sequence from the opposite direction to a forward hidden sequence and a backward hidden sequence. Due to it's effective contextual understanding ability, BiLSTM has been frequently used as a baseline classifier \citep{Moon2020BEEPKC}.

    
    \item Previous studies have shown that fine-tuning BERT-based
    models \citep{devlin2018bert, sanh2019distilbert, lewis2019bart, Liu2019RoBERTaAR}  yield impressive performance in various downstream tasks such as text categorization \citep{rogers2020primer, yamada2020luke, wu2019zero} question-answering \citep{Garg_Vu_Moschitti_2020,laskar2020contextualized}, summarization \citep{Liu2019TextSW, laskar2020query, laskar2020weakly, laskar2021domain}, sentiment analysis of social media posts \citep{moshkin2020application, tasnim2021being}, etc., since these models are pre-trained on a large amount of unlabeled data via leveraging self-supervised learning.
    
    
    \item Implementing a system that can detect the severity of depression from social media texts on devices with limited computational power may be difficult due to  the high parameter count of BERT (Base: 110 million). According to research on pre-trained models such as MegatronLM \citep{shoeybi2019megatron}, bigger models with several billions, if not trillions, of parameters usually result in superior performance on downstream tasks. However, the overall performance boost comes at the price of higher computational power and memory needs for both training and inference, rendering them unsuitable for use on the edge devices, such as smartphones. To address this issue, \cite{sanh2019distilbert} proposed DistilBERT, which has the same architecture as BERT and is pre-trained on the same corpus. By removing token-type embeddings and the pooler from the BERT implementation, DistilBERT reduces the number of layers by a factor of two, because hidden size dimensions have less of an influence on computation efficiency than the number of levels. DistilBERT is pre-trained through knowledge distillation via the supervision of a larger model incorporating triple loss functions (Distillation Loss - $L_{ce}$, Masked Language Modelling Loss - $L_{mlm}$, and Cosine Embedding Loss - $L_{cos}$). DistillBERT maintains 97\% of BERT performance on downstream tasks with 40\% fewer parameters. Additionally, it reduces the inference time of BERT in downstream tasks by around 60\%. The fundamental reason for this is a compression method called knowledge distillation, which enables a compact model to replicate the behavior of larger models as well as the components of triple loss.
\end{itemize}

Both BERT and DistilBERT relies on Auto Encoding (AE) language modeling during pre-training since the aim is to understand natural language representations. Although general transformer architecture proposed by \citet{vaswani2017attention} utilizes an encoder and a decoder network, BERT and DistilBERT, as pre-training models, only use the encoder to interpret the content of input sequences.

It is to be noted that, all of the baseline models that were chosen are data-driven approaches. As a result, these models are unable to extract semantic information from a context that is not explicitly in the data, unlike humans who can use their pre-existing knowledge to judge new contexts that they never encountered before \citep{garcez2019neural, cocarascu2018combining}. One solution to this problem could be the use of symbolic approaches. Unfortunately, these approaches fall short due to scalability. Recent approaches combine symbolic and data-driven approaches to solve this problem \citep{cocarascu2018combining, faghihi2021domiknows, steven2022modelling}. However, we limit ourselves to data-driven approaches to keep the baseline models simple.

\subsection{Classifier Configuration}
The training procedure of the baseline classifiers is demonstrated below, followed by the training parameters of the experiment.

\subsubsection{Support Vector Machine (SVM)}

SVM tries to draw a hyperplane that bests separates multi-dimensional data points in their potential classes and is ideal for binary classification \citep{Cortes2004SupportVectorN}. For multi-class classification, an `one-versus-one' approach with a Radial Basis Function (RBF) kernel was implemented. The values of the two crucial parameters for RBF kernel, \textit{C=0.5} and \textit{gamma=0.5}, were chosen based on several iterations of experiments. The entire dataset were split into 80\%-20\% partitions for training and testing the model. Several text pre-processing techniques, such as stopwords removal, bad symbols removal, text lower-casing, etc., were applied to both the training and testing data.

\subsubsection{Bidirectional LSTM (BiLSTM)}
\begin{figure}[h]
    \centering
    \includegraphics[width=\columnwidth]{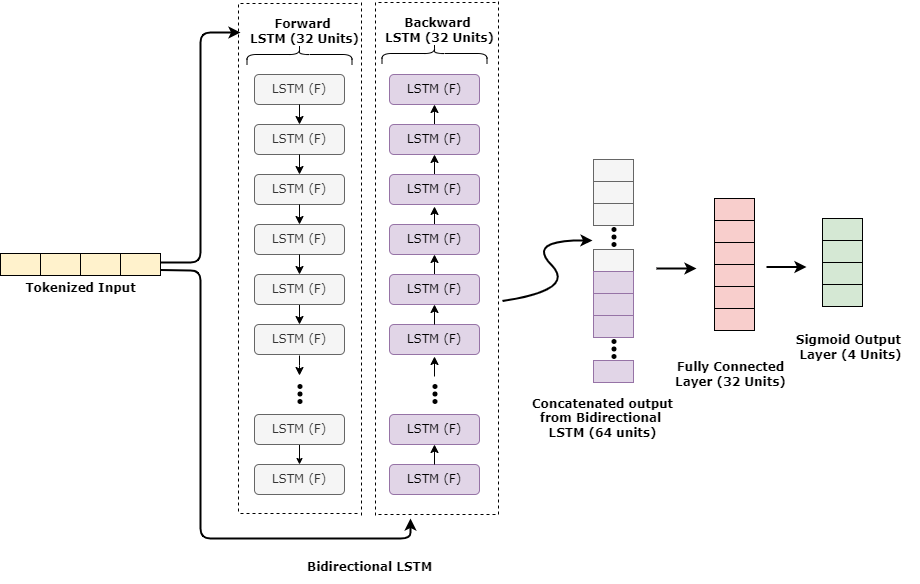}
    \caption{Architecture of BiLSTM Network}
    \label{fig:bilstm_architecture}
\end{figure}

BiLSTM can preserve the sequence information in both direction, backwards (right to left) or forward (left to right). To train the model, a bidirectional layer of 64 units were added after the word-embedding layer generated from the training data. The overall architecture of the BiLSTM network is illustrated in \figurename  \ref{fig:bilstm_architecture}. Similar text pre-processing techniques like SVM was deployed for BiLSTM as well. During the training phase, the hyperparameters for this experiment were fine-tuned using cross-validation, adopting 10\% of the data from the training samples as the validation set.

\subsubsection{Fine-tuning BERT \& DistilBERT}
Fine-tuning the pre-trained model weights in a task specific manner with respect to the tweet texts and their annotated labels is necessary to improve the classification performance considering that they are pre-trained using data from various sources.
\\
\textbf{Input Representation}\\
Before being fed into the pre-trained models for embedding, each tweet text were converted into an acceptable format. A single vector representing the entire input sentence is required to be passed to a classifier in order to complete the classification operation. BERT-based models use WordPiece tokenizer \citep{wu2016google}, which works by splitting the input sequence into full forms or  word pieces. In case of full form, a word is represented by one token string, whereas, for word pieces, a word is represented by multiple token strings. Using word pieces helps the models to identify related words as they share similar token strings, which is crucial for context understanding. Some special token strings are generated during tokenization to indicate the task type,
beginning of input sequence, mask, etc., e.g., 

\begin{itemize}
    \item `[SEP]' refers to the end of one input sequence and the beginning of another.
    \item `[CLS]' refers to the classification task.
    \item `[PAD]' is used to indicate the necessary padding.
    \item `[UNK]' stands for unknown token.
\end{itemize}

Classifiers used in this study require the input sequences to be of the same length, i.e., each tweet text should have an equal number of tokens after converting them to token strings. Since a maximum token length of 128 is used, if a comment contains less than 128 tokens, extra `[PAD]' tokens are added at the end of the token sequence. Both BERT and DistilBERT are pre-trained with 30K token vocabularies. So some new input data might appear while fine-tuning, which was not present in the pre-trained vocabulary. In that case, the new input substring is replaced by `[UNK]' token. Subsequently, the final input vector for the models was prepared by converting the token strings to integer token IDs. \\
\\
\textbf{Hyper-parameters Selection}\\
Fine-tuning and evaluating the classifiers required the proposed dataset to be splitted into three sets - train, validation, and test. Randomly selected 60\% tweets from each class were placed into the train set, and the rest of the tweets were equally distributed among the validation and test sets. Base-uncased\footnote{https://huggingface.co/bert-base-uncased} versions of the pre-trained models were implemented for fine-tuning with a total of 768 hidden output states.
Categorical Cross-Entropy loss function with AdamW optimizer \citep{loshchilov2017decoupled} was used that utilizes a fixed weight decay unlike common implementations of Adam optimizer \citep{kingma2014adam}. Considering that the learning rate was set to $3\times10^{\text{-}5}$ and 20\% of the steps were designated as warm-up steps, the training phase would use the first 20\% of the steps to raise the learning rate from $0$ to $3\times10^{\text{-}5}$. Here, steps denote the total number of times when the model weights get updated during the fine-tuning phase.

\begin{figure}[h]
    \centering
    \includegraphics[width=\columnwidth]{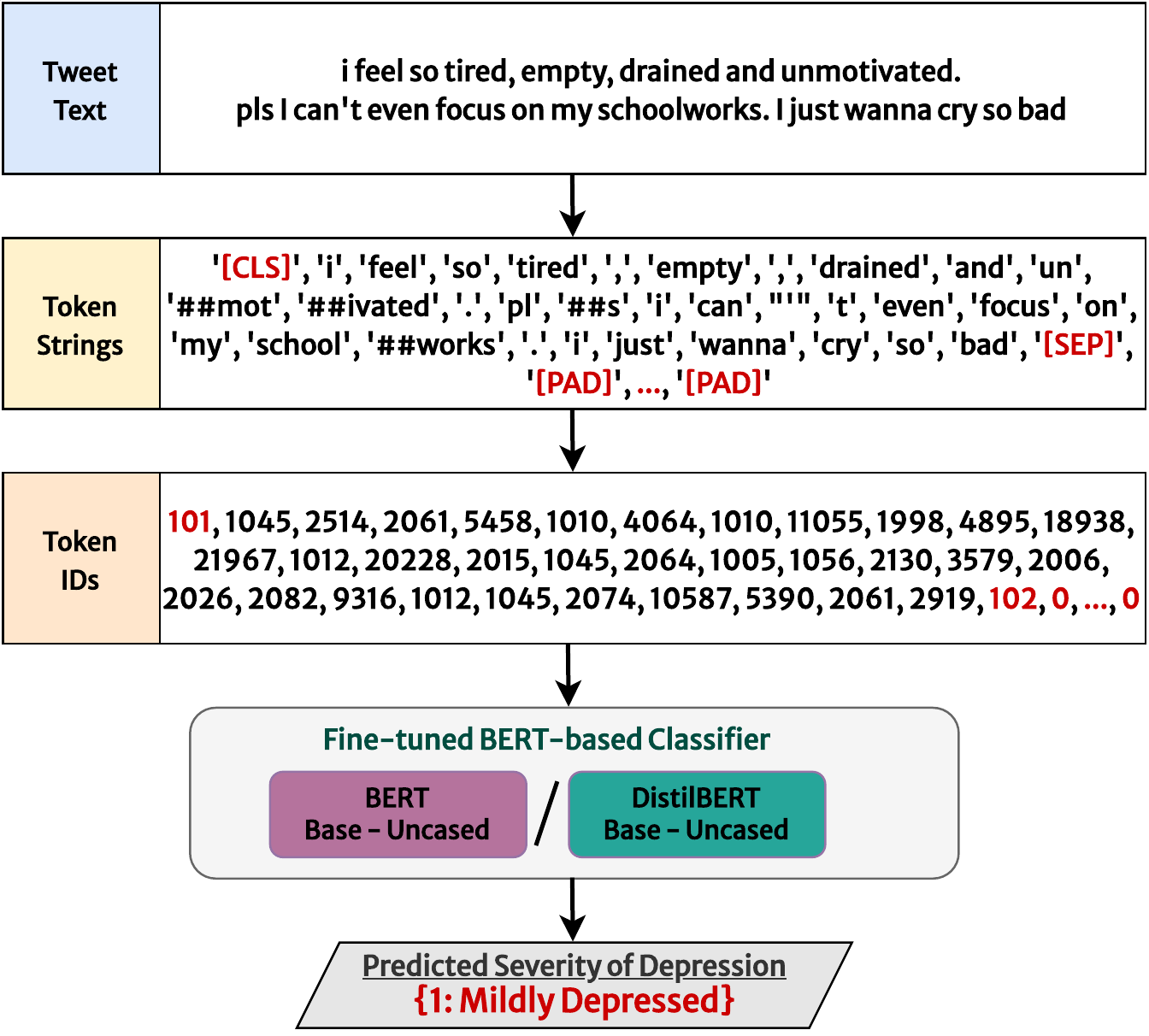}
    \caption{Severity of depression prediction from a sample tweet}
    \label{fig:architecture}
\end{figure}

Both of these models were fine-tuned in a supervised manner for 10 epochs with a training batch size of 16 on the proposed dataset to predict the severity of depression from tweets and achieved a good performance on all four classes. \figurename~\ref{fig:architecture} depicts the process of predicting the severity of depression using the fine-tuned classifiers from a sample tweet. 

\subsection{Evaluation Metrics}
Evaluation metrics play a crucial role in quantifying the performance of a predictive classifier \citep{sun2009classification}. Since the choice of metrics depends on the characteristic of the dataset, this can often lead to misleading conclusion regarding the experiment. For example, while evaluating an experiment on a highly imbalanced dataset, evaluation metrics such as accuracy, precision, or recall may lead to a conclusion that is practically useless. With imbalanced datasets, it is possible to reach very high accuracy without predicting any useful prediction since the majority predictions are from the densely populated classes \citep{leevy2018survey}. 

Other widely used evaluation metrics like precision, recall etc. have their own limitations. Precision is about exactness of classification task and relies only on true positive and false positive, it is possible to get a precision score of 1.0 by only one true positive prediction. On the other hand, recall is about completeness and depends solely on true positive and false negative. As a result, predicting all the samples as positive will give a recall of 1.0, whereas precision will be very low.

To tackle this issue, the Receiver Operating Characteristic (ROC) curve and area under the ROC curve (AUC-ROC) were used as evaluation measures in this work, such that models are evaluated based on how good they are at separating classes. ROC curve is a diagnostic diagram that calculates the False Positive 
Rate (FPR), and True Positive Rate (TPR) for a series of predictions made by the model at different thresholds to summarize the model's behavior which can be used to analyze the model's ability to discriminate classes.

In the ROC graph, each probability threshold is represented by a point, linked to form a curve. A model that with no discriminatory power between the classes will be represented by a diagonal line between fpr 0 and tpr 0 (co-ordinate: 0,0) to fpr 1 and tpr 1 (co-ordinate: 1,1). Points below this line reflect models with less competence than none. A flawless model will be represented as a point in the plot's upper left corner.

\section{Results and Discussions} \label{sec:result_discussion}
The performance of the baseline models on our dataset will be discussed in this section, followed by the potential unintended bias of this study.

\subsection{Classification Performance \& Analysis}

\begin{table}[t]
\centering
\caption{Performance Comparison of Baseline Models}
\label{tab:results}
\begin{tabular}{l l C{2cm}} 
\toprule
\textbf{Model} & \textbf{Class Name} & \textbf{ROC AUC Score}\\
\midrule
\multirow{4}{*}{SVM}       & Non-depressed       & 0.514816                                                                       \\ 
                            & Mild                & 0.511343                                                                     \\ 

                            & Moderate            & 0.512785                                                                    \\ 
                            & Severe              & 0.547684                                                                      \\ 
\midrule
\multirow{4}{*}{BiLSTM} & Non-depressed       & 0.692522                                                                       \\ 
                            & Mild                & 0.565517                                                                    \\ 
                            & Moderate            & 0.795351                                                                      \\ 
                            & Severe              & 0.755356                                                                      \\
\midrule
\multirow{4}{*}{BERT}       & Non-depressed       & 0.763699                                                                       \\ 
                            & Mild                & 0.740019                                                                       \\ 

                            & Moderate            & 0.748115                                                                       \\ 
                            & Severe              & 0.826488                                                                       \\ 
\midrule
\multirow{4}{*}{DistilBERT} & Non-depressed       & 0.788841                                                                       \\ 
                            & Mild                & 0.747211                                                                       \\ 
                            & Moderate            & 0.787959                                                                       \\ 
                            & Severe              & 0.866003                                                                       \\
\bottomrule
\end{tabular}

\end{table}


\begin{figure*}[h]
\centering
\begin{subfigure}{\columnwidth}
  \centering
  \includegraphics[width=\linewidth]{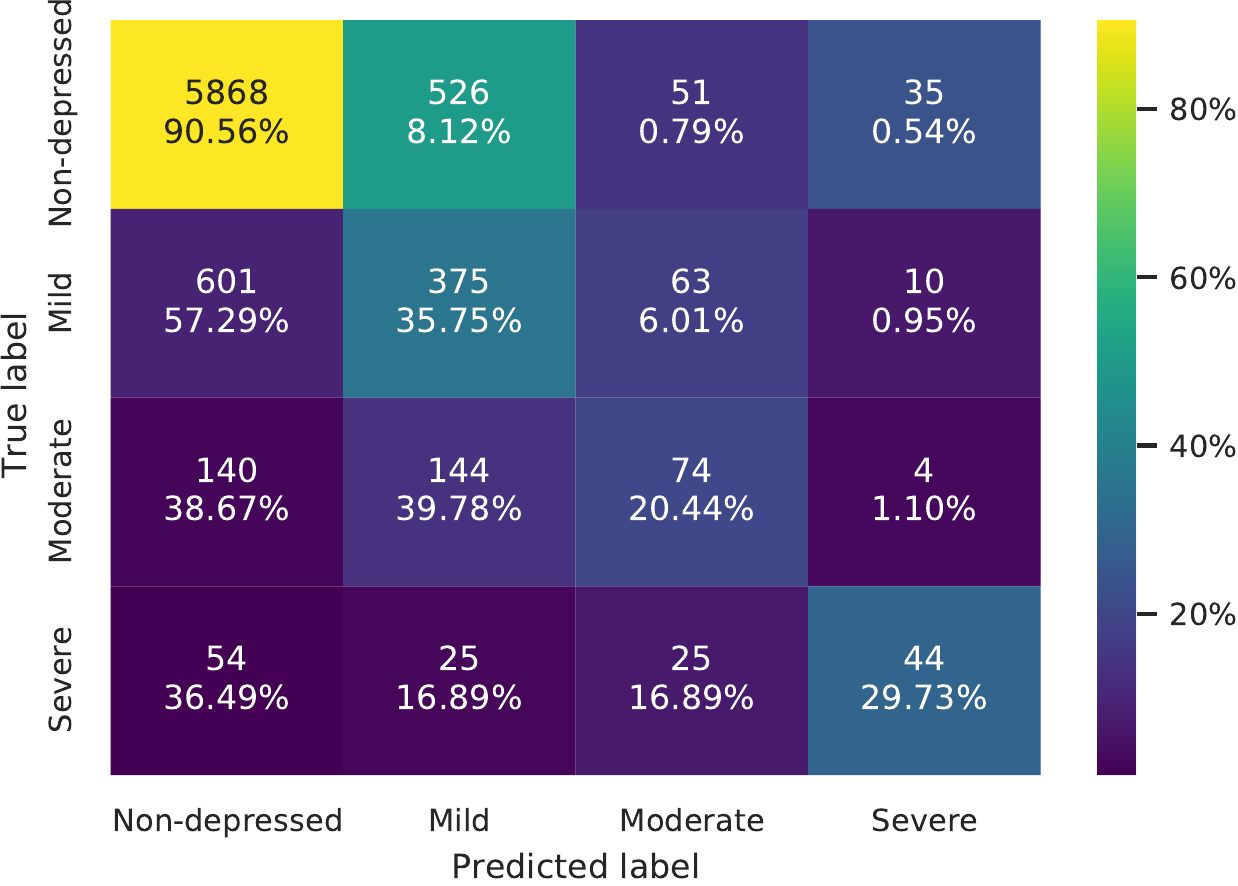}
  \caption{BERT}
  \label{fig:conf_sub1}
\end{subfigure}%
\begin{subfigure}{\columnwidth}
  \centering
  \includegraphics[width=\linewidth]{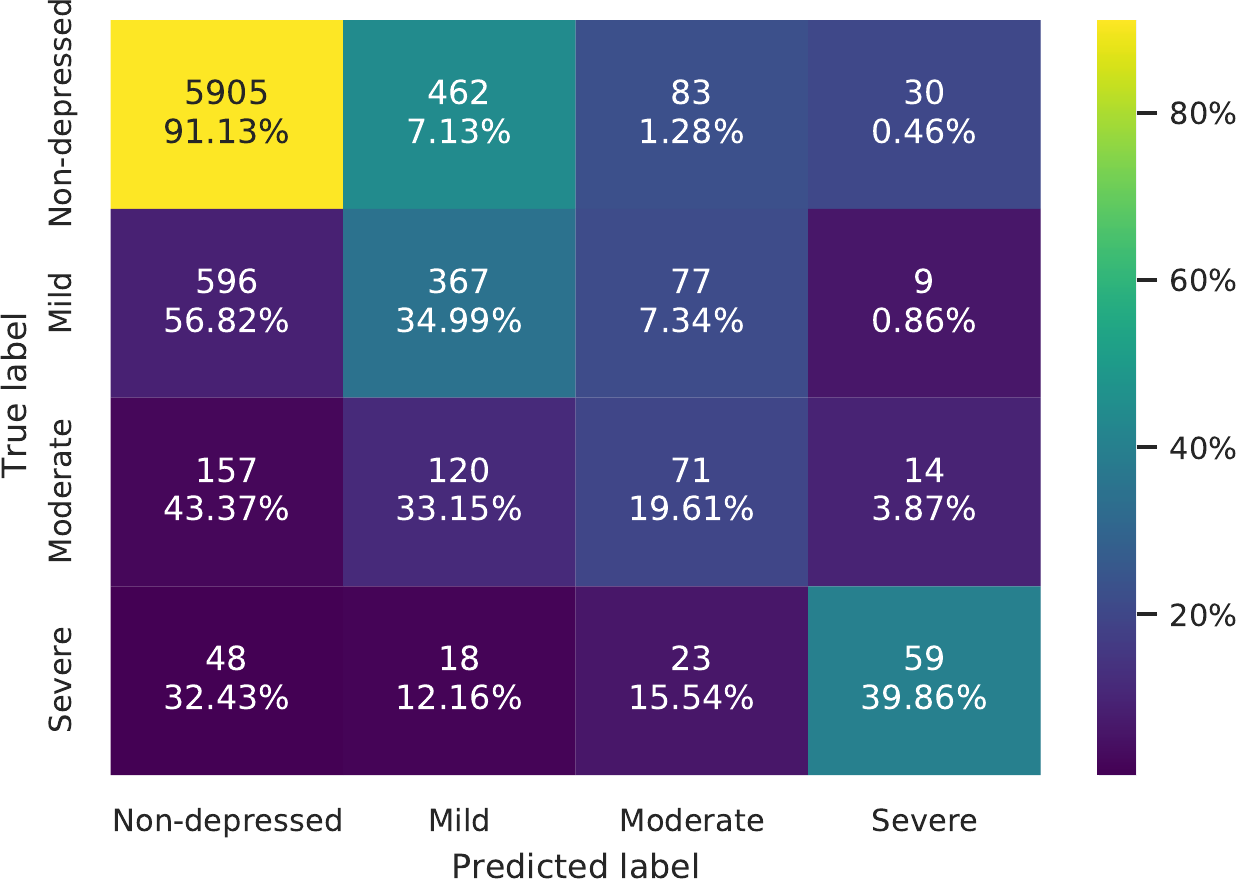}
  \caption{DistilBERT}
  \label{fig:conf_sub2}
\end{subfigure}
\caption{Confusion matrix obtained by evaluating test set using fine-tuned classifiers}
\label{fig:conf_mat}
\end{figure*}



\begin{table*}[t]
    \centering
    \caption{Model Predictions for the Terminal Classes}
    \label{tab:error}
    \begin{tabular}{L{8.5cm} L{2.5cm} L{2.5cm}}
        \toprule
        \textbf{Tweet Text} & \textbf{Ground Truth} & \textbf{Predicted Label}\\
        \toprule
        I knew self destruction ain’t the only way… & \textit{non-depressed} & \textit{severe} \\
        Yes actually. I feel like it invalidates what queer people go through when they're depressed and attempt/want to attempt commit suicide. & \textit{non-depressed} & \textit{severe} \\
        my stomach is killing me. my whole body hurts i’m so exhausted & \textit{non-depressed} & \textit{severe} \\
        \midrule
        i inherited a thirst for self destruction and i’m scared of it & \textit{severe} & \textit{non-depressed}\\
        Sorry I know what this feels like lost 23 of my best friends in combat… as well as suicide coming back home… depression does suck, but we can do this & \textit{severe} & \textit{non-depressed}\\
        I don’t like to brag. BUT, I don’t think there’s a soul on this earth that does self destruction like I do. & \textit{severe} & \textit{non-depressed}\\

        \bottomrule
    \end{tabular}
    
\end{table*}


According to the results shown in \tablename~\ref{tab:results}, it can be observed that SVM and BiLSTM were outperformed by the two transformer based models by a large margin. Transformer-based models that are used in this study can learn each word's context from the words that appear before and after it and are also pre-trained on a large corpus. Since effective context understanding from the input representations is very crucial to the task of severity detection from tweets, these models are likely to outperform traditional deep learning based models such as LSTM, BiLSTM or unidirectional transformer based models such as OpenAI GPT \citep{radford2018improving} where each token is capable of managing only the preceding tokens in the transformer's self-attention layers. As BiLSTM can also learn contexts of words in both directions, it seems to achieve decent performance in some classes as well.

It can also be observed that DistilBERT outperformed BERT in all classes. Since DistilBERT is pre-trained under the supervision of its parent model, BERT through knowledge distillation, it is able to preserve 95\% performance of the base uncased BERT \citep{sanh2019distilbert} which is divergent to the experimental results shown in this study. The experiments were conducted in a computationally limited environment with a comparatively smaller batch size and fine-tuned only for 10 epochs. It is likely that, BERT will outperform DistilBERT if the models are fine-tuned for higher number of iterations with further hyper-parameter tuning.

As seen from \tablename~\ref{tab:dataset_class_dist}, the proposed dataset is mostly comprised of the samples from the \textit{`non-depressed'} class, in which both models showed commendable performance in detecting classes with relatively smaller number of samples for other classes as well. From the confusion matrices in \figurename~\ref{fig:conf_mat}, it can also be noticed that both the models performed better on the two terminal classes \textit{`non-depressed'} and \textit{`severe'} than the two closely related classes, \textit{`mild'} and \textit{`moderate'}. Upon careful observation, it was found that wrong predictions of the samples were mostly due to models failing to comprehend the contextual meaning of the comments properly and instead generalizing based on specific keywords to predict the final label. For example, as shown in Table \ref{tab:error}, in few cases where the ground truth is \textit{`non-depressed'} but the predicted label by the models is \textit{`severe'} and vice-versa, most of these cases contain words related to suicide, depression, self-destruction, self-harm, etc. So, this enables the room for further improvement through error analysis.

\begin{figure*}[h]
\centering
\begin{subfigure}{\columnwidth}
  \centering
  \includegraphics[width=\linewidth]{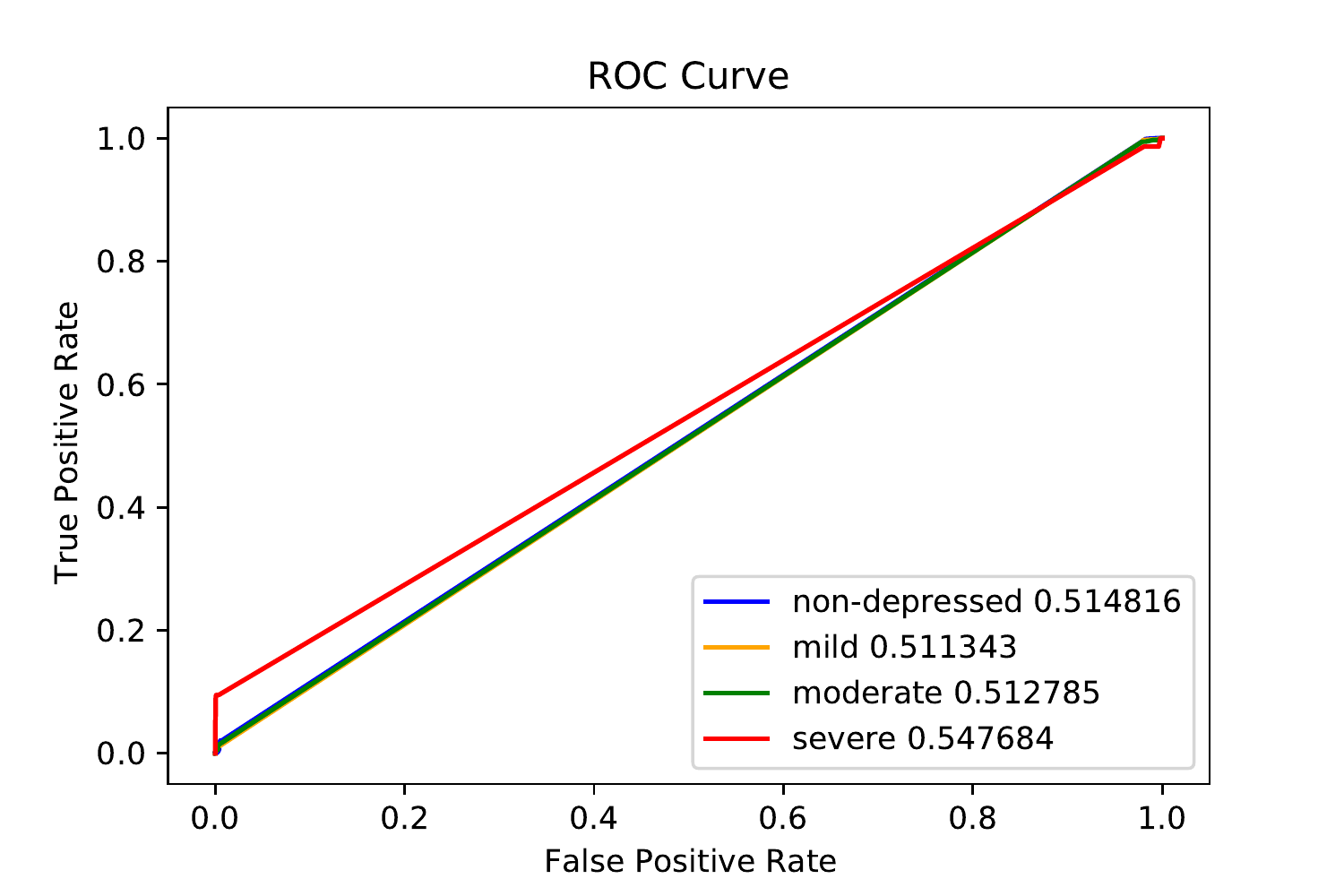}
  \caption{SVM}
  \label{fig:roc_sub4}
\end{subfigure}
\begin{subfigure}{\columnwidth}
  \centering
  \includegraphics[width=\linewidth]{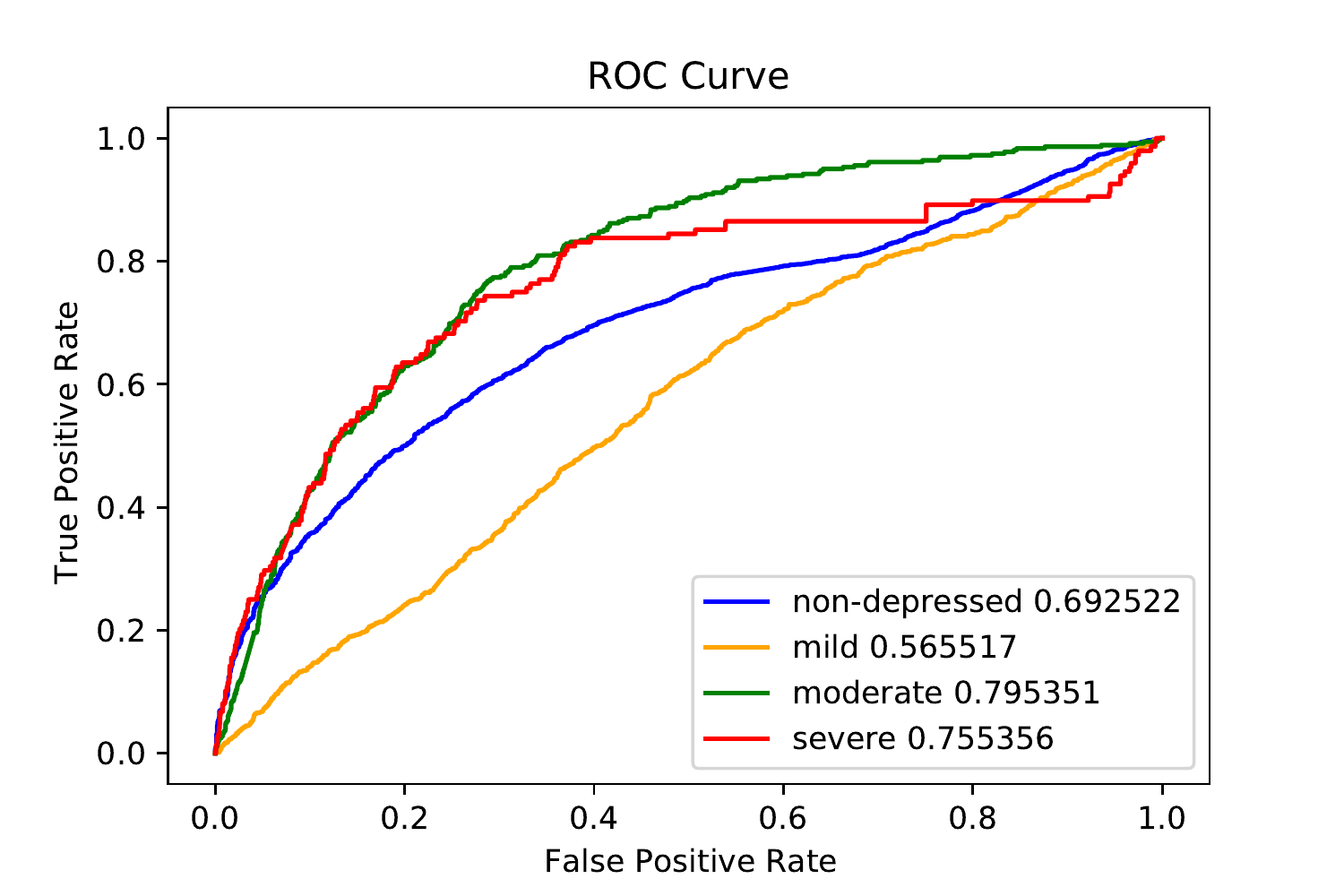}
  \caption{BiLSTM}
  \label{fig:roc_sub3}
\end{subfigure}
\begin{subfigure}{\columnwidth}
  \centering
  \includegraphics[width=\linewidth]{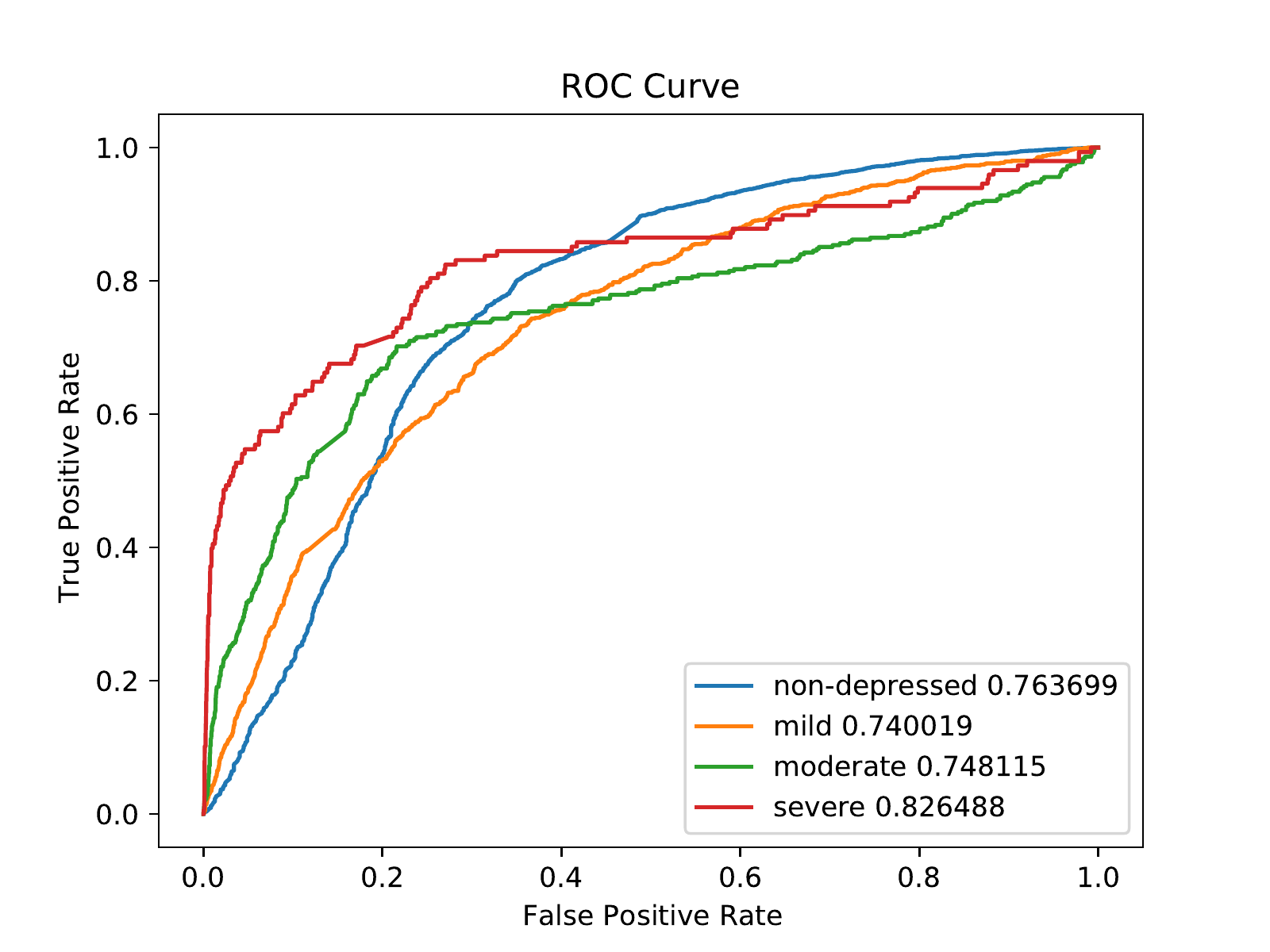}
  \caption{BERT}
  \label{fig:roc_sub1}
\end{subfigure}
\begin{subfigure}{\columnwidth}
  \centering
  \includegraphics[width=\linewidth]{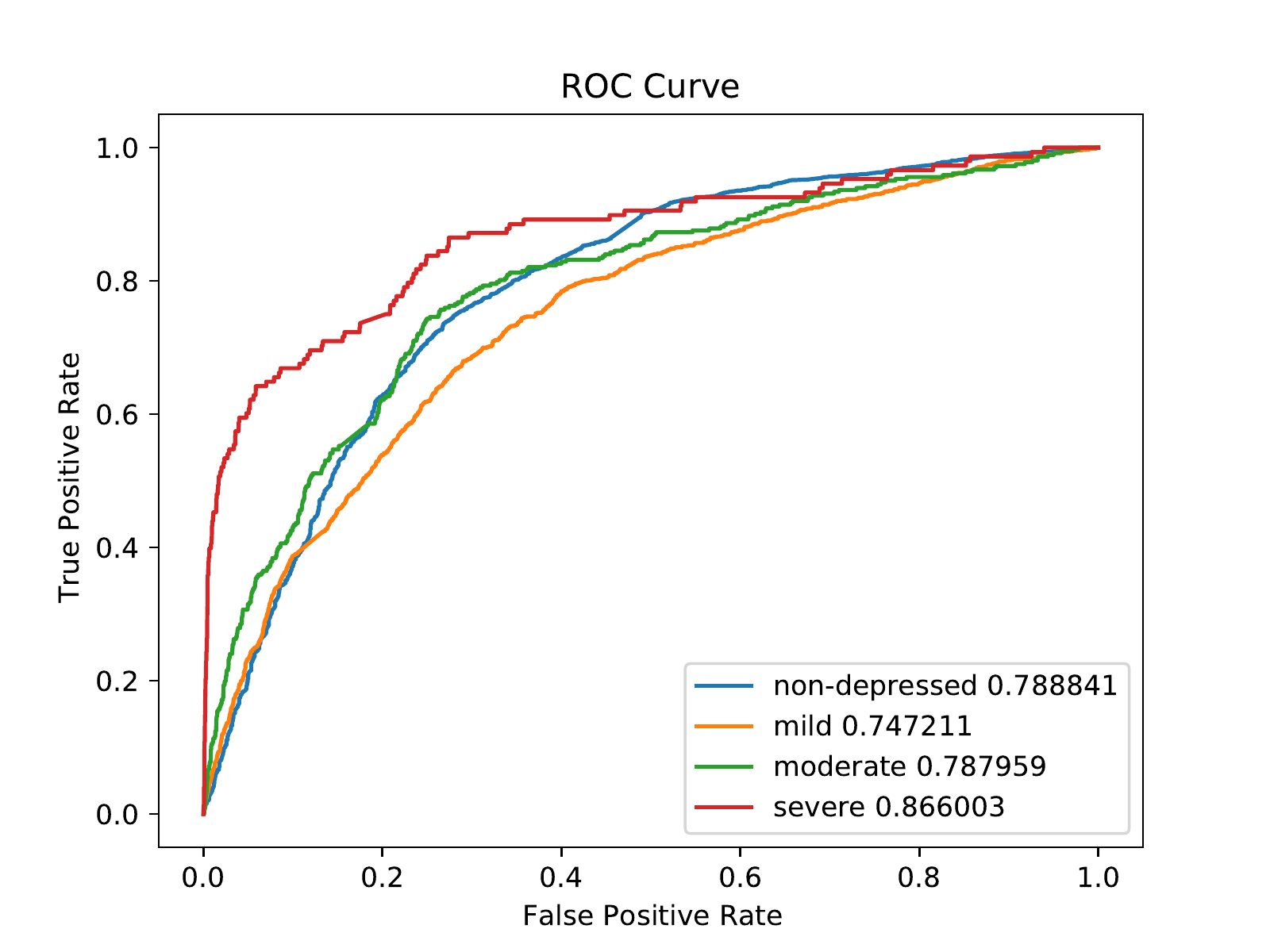}
  \caption{DistilBERT}
  \label{fig:roc_sub2}
\end{subfigure}
\caption{Class-wise AUC-ROC curves}
\label{fig:roc}
\end{figure*}

For the proposed dataset, ROC curves using the test predictions from the baseline classifiers is presented in \figurename~\ref{fig:roc}. These plots are summarized by calculating the area under the ROC curve (AUC-ROC) in \tablename~\ref{tab:results}. The better performance of DistilBERT and BERT are also distinguishable from the class-wise AUC-ROC curves in \figurename~\ref{fig:roc}.

\subsection{Potential Unintended Bias}\label{sec:potentialUnintended}

\figurename~\ref{fig:kde} shows that \textit{non-depressed} and \textit{severe} classes are more condensed towards the complete agreement of the annotators. As these two classes lie on the two polarities and have distinguishable attributes, the annotators were likely to agree more on these two class labels while annotating. The main challenge was to differentiate between the other two classes, i.e., \textit{moderate} and \textit{severe} for their inherent subtleties and congruent attributes. With the tweet corpus being in English, and considering the subtle attributes of the different severities of depression, the dataset was likely to achieve higher annotation quality if the annotation was done by annotators with first-language proficiency in English. As the study requires a large pool of annotators and demands consistent supervision and interaction of the annotators with the collaborator psychologists, it limits the choice of recruiting only English-speaking annotators. This was attempted to be reduced by recruiting annotators with excellent abilities in English and pre-screening was done before the final pool of annotators were selected.

Another challenge that appeared in a similar context for the annotators was to avoid their individual bias while deciding the class labels. The source of the tweets and their nuances in attributes complicated the annotation task and potentially introduced bias into the dataset. From the manual inspection of the scraped tweet samples, it was observed that the majority of the samples were from the North American region, while all the annotators were from South Asia. This can introduce a clear cultural and geographic bias in the annotation procedure. Though the tweets were presented in isolation to the annotators, without all the related information (i.e., tweet ID, retweets, location, etc.) and without the surrounding context of scraping the tweets, the collaborator psychologists speculated a bias in the annotation as there is a clear cultural and expressional difference between the users and annotators of the tweets. The annotators were reminded several times throughout the annotation process to avoid their personal bias and strictly follow the guidelines laid out by the psychologists, which included a document containing high-level descriptions of the attributes of the classes. This issue of systematic bias is common for large datasets, as addressed by \citet{Vidgen2019ChallengesAF}, especially for complex multi-class tasks of this kind.

The data extension tool used for this study is Wordnet, which was initially released in the mid 1980s. Though it has been updated over time, due to the continuous evolution of language, people today often use a vocabulary on social media that can differ significantly from the one that Wordnet represents. Moreover, some of the semantic relations enlisted in wordnet  are more suited to concrete notions than to abstract ones \citep{Rudnicka2018LexicalPO}. For example, it is easy to create hyponyms/hypernym relationships to illustrate that a ``Pinaceae'' is a type of ``tree'', a ``tree'' is a type of ``plant'', and a ``plant'' is a type of ``organism'', but it is difficult to classify emotions like ``anxiety'' or ``delight'' into equally deep and well-defined associations. Finding appropriate seed terms that best capture the depressive emotion of people on social media might be substantially hindered by these limitations.

\section{Conclusions and Future Work}\label{sec:conclusionsAnd}
This work introduced a new typology for diagnosing depression severities from social media texts, as well as a unique dataset of labeled tweets with a confidence score for each label. The dataset was constructed based on strong ground truths and clinical validation, and it is expected to help alleviate the scarcity of mental health data to some extent. The description of the process and challenges in creating such a dataset may motivate researchers to collect similar corpora of this scale from other social media and discussion forums. The experimental results indicated that existing state-of-the-art models often fail to understand the contextual undertone of the data samples. Developing a model that is capable of comprehending the subdued relationship and differences among the depression severities can result in an even better understanding of human cognition. Moreover, analysis of the classification performance indicates that there is no distinct division of keywords among different depression severities. Same keyword might be used differently to express different emotions, rather it is more important to understand the context of the tweet to diagnose the severity of depression. Broader implications of this research may include personalizing and directing preventative and awareness messages by health professionals to the users in need.

The seed terms for each symptoms of PHQ-9 in this study was extended by Wordnet \citep{miller1995wordnet}. Considering the fast evolving nature of languages in social media, future studies can utilize more recent lexical databases with larger semantic network to extend the seed terms. For example, CMU pronouncing dictionary \footnote{http://www.speech.cs.cmu.edu/cgi-bin/cmudict}, MRC Psycholinguistic Database \citep{Coltheart1981TheMP}, The Verb Semantics Ontology Project \citep{Fukushima1984OntologyIT} are other available lexical databases that can be used in seed term extension. Additionally, authors can also develop their own domain-specific lexical database by vector proximity using a domain-specific corpus as a starting point. These approaches can build a keyword list that better extracts depression related symptoms posted on social media nowadays. The baseline classification result of the dataset was provided by fine tuning two modern pre-trained models, namely BERT and DistilBERT. It is worth noting that several features in the dataset, such as \textit{replies\_count} and \textit{retweets\_count}, were not used during training, and no pre-processing was performed on the data. Therefore, more accurate classification might be achieved on this dataset by: (1) including a pre-processing technique to clean the data before training, (2) increasing trainable instances by augmentation to eliminate the class imbalance of the dataset, (3) utilizing other features of the dataset during training, (4) fine-tuning more robust pre-trained models, etc. Because the data was collected during the post COVID-19 pandemic phase, careful examination of the dataset can provide valuable insight into the impact of the pandemic on people's mental health. Moreover, the DEPTWEET dataset can be expanded by annotating the remaining 2510 data samples for which a class label could not be determined due to annotators' disagreement. Further work may also include refining the annotation task by including annotators from similar cultural and geographic contexts and exploring the unintended biases in the data and model.




\printcredits

\bibliographystyle{cas-model2-names}

\bibliography{cas-refs.bib}

\end{document}